\pdfoutput=1
\documentclass[11pt]{article}

\usepackage{EMNLP2023}

\usepackage{times}
\usepackage{latexsym}
\usepackage{booktabs}
\usepackage{graphicx}
\usepackage[T1]{fontenc}
\usepackage[utf8]{inputenc}

\usepackage{microtype}

\usepackage{inconsolata}

\usepackage{xcolor}
\usepackage{setspace}

\usepackage[symbol]{footmisc}

\title{\textsc{Tweet Insights}: A Visualization Platform \\ to Extract Temporal Insights from Twitter}
\author{\textbf{Daniel Loureiro$^1$, Kiamehr Rezaee$^1$, Talayeh Riahi$^1$, Francesco Barbieri$^2$,}\\ \textbf{Leonardo Neves$^2$, Luis Espinosa Anke$^{1,3}$, Jose Camacho-Collados$^1$} \\
$^1$ Cardiff NLP, School of Computer Science and Informatics, Cardiff University, UK  \\
$^2$ Snap Inc., Santa Monica, California, USA 
$^3$ AMPLYFI \\ 
\texttt{cardiffnlp.contact@gmail.com}
}

\begin{document}

\maketitle

\begin{spacing}{1.0}

\begin{abstract}
% In this paper, we present a collection of time series derived from the Twitter API and language models fine-tuned for the Twitter domain. These time series, covering the past 5 years, capture changes in n-gram frequencies, word meaning, emotional valence, and topic prevalence. This tool can be used to better detect and characterize meaning shift, particularly by disentangling popularity peaks from other dimensions such as diachronic similarity and topic distribution. We also provide online demo for quick experimentation and explore different use cases in this paper.

This paper introduces a large collection of time series data derived from Twitter, postprocessed using word embedding techniques, as well as specialized fine-tuned language models. This data comprises the past five years and captures changes in n-gram frequency, similarity, sentiment and topic distribution. The interface built on top of this data enables temporal analysis for detecting and characterizing shifts in meaning, including complementary information to trending metrics, such as sentiment and topic association over time. We release an online demo for easy experimentation, and we share code and the underlying aggregated data for future work. In this paper, we also discuss three case studies unlocked thanks to our platform, showcasing its potential for temporal linguistic analysis.
\end{abstract}

%language models fine-tuned for the domain

\section{Introduction}

Social media platforms are fundamental tools in today's society, and Twitter, in particular, has emerged as a thermometer for modern communication. While there have been numerous research studies targeting Twitter for societal investigations \cite{zhuravskaya2020political,wohn2011tweeting,lim2015social}, 
%partly due to the ease of access compared to other social media platforms, 
analysing it at a large scale, and for a prolonged period is not an easy task. This requires both a compilation of a large general-purpose corpus, as well as the usage of cutting-edge NLP techniques for data analysis. Moreover, the noisiness and informal nature of the platform make this task even especially challenging \cite{baldwin2013noisy,morgan2014information}. 

From a linguistic perspective, social media offer a valuable corpus for analysing word meaning change over time \cite{del-tredici-etal-2019-short,loureiro-etal-2022-tempowic}, alongside tracking changes related to word usage. Several factors can influence language and topic discussions on Twitter, such as the release of a new movie (e.g., \textit{Parasite}), the emergence of a new meaning for a certain word (e.g., \textit{delta}), or an event that causes a sudden change in the sentiment toward a particular entity (e.g., \textit{Alec Baldwin}).

In this paper, we present \textsc{Tweet Insights}, a platform for analysing Twitter textual data from a temporal perspective. Tweet Insights is released as a set of pre-computed time series and an online demo\footnote{A short video showcasing the online demo is available at \url{https://tinyurl.com/ycxwh9mu}} for analysing historical multi-faceted word-based statistics in real time. With Tweet Insights, non-expert users can gather temporal statistics for word frequency, embedding similarity, sentiment, and topics. 
%Time series - meaning shift - NLP techniques - Google trends/ngrams? - 
%\url{https://tinyurl.com/ycxwh9mu}}

\section{Related Work}

%Twitter has long been targeted for several studies on Computational Social Science.
Numerous studies have focused on using Twitter as a data source in Computational Social Science research, including statistical tools helping in this regard \cite{o2010tweetmotif}.
Recently, \citet{https://doi.org/10.48550/arxiv.2301.11429} analysed all tweets posted on the social platform during a particular day (375M), with the goal of better understanding bot activity and user demographics, among other aspects.
Most similar to our work, Storywrangler \cite{doi:10.1126/sciadv.abe6534} continuously collects a large sample of tweets and provides several visualizations related to word frequency through an online interface.

While there are commercial tools in the intersection between advanced social media monitoring, decision-making and market intelligence (e.g., Bloomberg Terminal \cite{dredze2016twitter}, FactSet or Thomson Reuters Eikon), these applications are often circumscribed to the financial domain, and most importantly, are not open-source and have a high cost\footnote{A Bloomberg Terminal license, for example, costs around USD 25,000.}.  To our knowledge, Hedonometer \cite{doi:10.1073/pnas.1411678112} is the only non-commercial documented tool analysing aspects complementary to popularity at scale, namely, tracking positive sentiment on the platform.
In terms of language evolution, the research community has notably used Twitter to analyse meaning shift using distributional methods \cite{Mitra2014ThatsSD,10.1145/2736277.2741627}, although at a smaller scale than the previously mentioned related works.
As such, our work is the first attempt at deriving deeper insights at scale from this social platform, leveraging state-of-the-art language models and embedding methods to provide a more holistic understanding beyond surface-level count-based metrics.

%Works such as \citet{doi:10.1126/sciadv.abe6534} and 
%Google Ngrams \cite{}
%Storywrangler \cite{doi:10.1126/sciadv.abe6534}.
%Day on Twitter \cite{https://doi.org/10.48550/arxiv.2301.11429}.
%Diachronic word embeddings and semantic shifts: a survey \cite{Kutuzov2018DiachronicWE}
%That’s sick dude \cite{Mitra2014ThatsSD}
%Characterizing the Google Books Corpus: Strong Limits to Inferences of Socio-Cultural and Linguistic Evolution \cite{Pechenick2015CharacterizingTG}
%The Changing Psychology of Culture From 1800 Through 2000 \cite{Greenfield2013TheCP}

\section{\textsc{Tweet Insights} Development: Data and Models}
%TOREMOVE

This work proposes a set of four time series designed for exploring meaning shifts through different facets, among other potential applications.
In this section, we describe the data and methods used to derive each of them. %TOREMOVE

\subsection{Corpus}

We use the Twitter Academic API to collect a total of 220M English tweets, covering the period between the start of 2018 and the end of 2022.
We follow the process described in \citet{loureiro-etal-2022-timelms} to compile a representative sample of tweets using the Twitter API -- ignoring retweets, media tweets, and requiring stopwords in the tweet's text to increase the likelihood of sampling tweets with meaningful content.
This process results in a collection of tweets distributed uniformly across different time scales (month, day, and hour), with the exception that more recent months sample additional intervals 
(see \autoref{sec:appendix} for more details). %We also follow the preprocessing described in \citet{loureiro-etal-2022-timelms} 
Tweets are preprocessed by anonymizing non-verified\footnote{Only `legacy' verified users, collected on 2022-11-09.} user mentions, and removing URLs.

\paragraph{Word selection.} Once the corpus is compiled, our goal is to select a representative set of words from the corpus. To this end, we tokenize tweets using NLTK's TweetTokenizer \cite{10.5555/1717171}, and recognize n-grams using the statistical scoring method proposed in \citet{NIPS2013_9aa42b31}, through the implementation available from \citet[Gensim v4.1.2]{rehurek_lrec}\footnote{Recognizing bigrams and trigrams with parameters: max\_vocab\_size=100M, min\_count=10, threshold=10, ignoring connector words.}.
For methods that rely on co-occurrence metrics, namely n-gram recognition (and associated frequencies), and learning diachronic word embeddings, we use lowercased texts.
See \autoref{tab:stats} for more details about corpus size and a breakdown by token types. %Text classification methods are applied using case-sensitive models and texts.
While casing is preserved for classification methods, for improved consistency across time series, and given the informal nature of Twitter, resulting statistics are aggregated for lowercased n-grams.

\begin{table}[]
\centering
\begin{tabular}{@{}lr@{}} \toprule
Tweets & 220,829,111 (220M)\\
Tokens & 4,186,915,103 (4B) \\ \midrule
\multicolumn{2}{c}{Token Types} \\ \toprule
Unigrams & 321,073 \\
Bigrams & 123,604 \\
Trigrams & 12,946 \\
Hashtags & 170,138 \\
Usernames & 96,639 \\ \bottomrule
\end{tabular}
\caption{Corpus statistics and token type counts (unique). Note that bigrams and trigrams are recognized while ignoring stopwords (e.g. \textit{game of thrones} would be counted as a bigram). Therefore some words may contain more than three actual tokens.}
\label{tab:stats}
\end{table}

\paragraph{Language models.}

As a byproduct of our demo, we also release two RoBERTa language models (base\footnote{\url{https://huggingface.co/cardiffnlp/twitter-roberta-base-2022-154m}} and large\footnote{\url{https://huggingface.co/cardiffnlp/twitter-roberta-large-2022-154m}}) trained on the collected Twitter API data.
These models are trained on a sample of 154M tweets covering the periods between 2018-01 and 2022-12.
This sample results from filtering our corpus of 220M tweets, following the preprocessing procedure described in \citet{loureiro-etal-2022-timelms}, which removes near-duplicates and tweets from the top most popular 1\% users (i.e., potential bots).
For training, we followed the same procedure of \citet{barbieri2020tweeteval} by training the original checkpoints until converging on an validation set of 50k tweets.

\subsection{Time series}

Once the reference corpus is compiled, we split the data into monthly intervals, starting from January 2020. Data for the months between 2018-01 and 2019-12 is aggregated into a single period (i.e., `up to 2020'), to establish prior baselines.
While in Section \ref{sec:demo} we provide details of our main online visualization platform, we also provide an easy-to-use Colab notebook\footnote{\url{https://tinyurl.com/8pvb7wvb}} with code to access the aggregated time series data in a programmatic way and create custom data visualizations.
%\jose{While in Section \ref{sec:demo} we provide details of our main online visualization platform, in the supplementary material we also include instructions and easy-to-use Colab notebooks with code to access the aggregated time series data in a programmatic way and create data visualizations.}
In the following we describe the four main aggregated time series available in \textsc{Tweet Insights}, namely frequency, embedding distance, sentiment and topic distribution. %TOREMOVE

\subsubsection{Frequency}
The frequency time series is based on n-gram counts obtained from the full collection of 220M tweets. The frequency of each word in the vocabulary is aggregated monthly.
Bigram and trigram frequencies add to their respective unigram components (e.g., an occurrence for \textit{Joe Biden} also increments counts for \textit{Joe} and \textit{Biden}). Our resource includes both absolute and normalized (words per million) counts. 

%\subsubsection{Similarity (Distance)}
\subsubsection{Embedding distance}
In order to compare embedding similarity for the same n-gram at different time periods, we employ the TWEC method proposed by 
\citet{Di_Carlo_Bianchi_Palmonari_2019} to efficiently learn diachronic word embeddings\footnote{Based on word2vec embeddings (300-d, min\_count=10).} for every month covered in our corpus.
As done for training the RoBERTa language models (described in 3.2), this step also applies the same preprocessing to improve the resulting embeddings.
Considering that TWEC starts by learning a so-called `compass' set of embeddings from the concatenation of text from all periods, we undersample more recent months (which have additional data) in order to learn compass embeddings from a corpus that is balanced across monthly intervals.
%Additionally, for this step, we preprocess tweets to remove near-duplicates and tweets from the top 1\% users (i.e., potential bots), following \citet{loureiro-etal-2022-timelms}.
The atemporal compass embeddings are then used to initialize embeddings for particular time periods, which are tuned for their periods by re-training on data corresponding to their temporal interval.
%These embeddings are trained on the same 154M preprocessed corpus used to train the RoBERTa language models (described in X).

%\jose{\paragraph{Word embeddings.} TODO: Explain word embeddings and add link}

The time series resulting from these embeddings %\footnote{Available here: \jose{URL}} 
features the cosine distance between the embedding from the first month with available data (the embedding prior, corresponding to `up to 2020' for most n-grams), and embeddings corresponding to each subsequent month. Consequently, there are no cosine distances to report for the first months with available data.

\subsubsection{Sentiment}
%rewrote as it was a bit verbose
%  We use a recent TimeLMs \cite{loureiro-etal-2022-timelms} RoBERTa-based model\footnote{\url{https://huggingface.co/cardiffnlp/twitter-roberta-base-sep2022}}, fine-tuned for sentiment classification, to obtain negative, neutral, and negative scores for every tweet in our corpus. This model has been proved to be reliable for sentiment analysis, 73.7 according to the official macro-averaged recall metric on the SemEval-2017 sentiment analysis task \cite{camacho-collados-etal-2022-tweetnlp}, while being of a reasonable size to run millions of inferences.
% Since our goal is to provide the most accurate predictions, we use a model fine-tuned on all splits of the SemEval-2017 sentiment classification task \cite{rosenthal-etal-2017-semeval}.
We use the Twitter-trained RoBERTa-base language model %\cite{loureiro-etal-2022-timelms}  
%model\footnote{\url{https://huggingface.co/cardiffnlp/twitter-roberta-base-sep2022}} , 
and fine-tune it for sentiment classification to obtain negative, neutral, and negative scores for every tweet in our corpus. Taking the SemEval-2017 sentiment analysis dataset as a reference \cite{rosenthal-etal-2017-semeval}, the model achieves 73.7 macro-averaged recall and outperforms other general-purpose models \cite{camacho-collados-etal-2022-tweetnlp}, while being of a reasonable size to run millions of inferences. Since our goal is to provide the most accurate predictions, for inference we use a model fine-tuned on all splits of the SemEval dataset.

With predictions for all tweets, we compile our sentiment time series as the mean prediction scores for tweets where a particular n-gram occurs.
To make comparisons more reliable across monthly intervals, for each n-gram, we set a minimum threshold of 10 tweets per month and compute the mean scores considering a sample of up to 1,024 tweets per month (depending on availability).

\begin{figure*}[!htpb]
    \centering
    \includegraphics[width=1.0\linewidth]{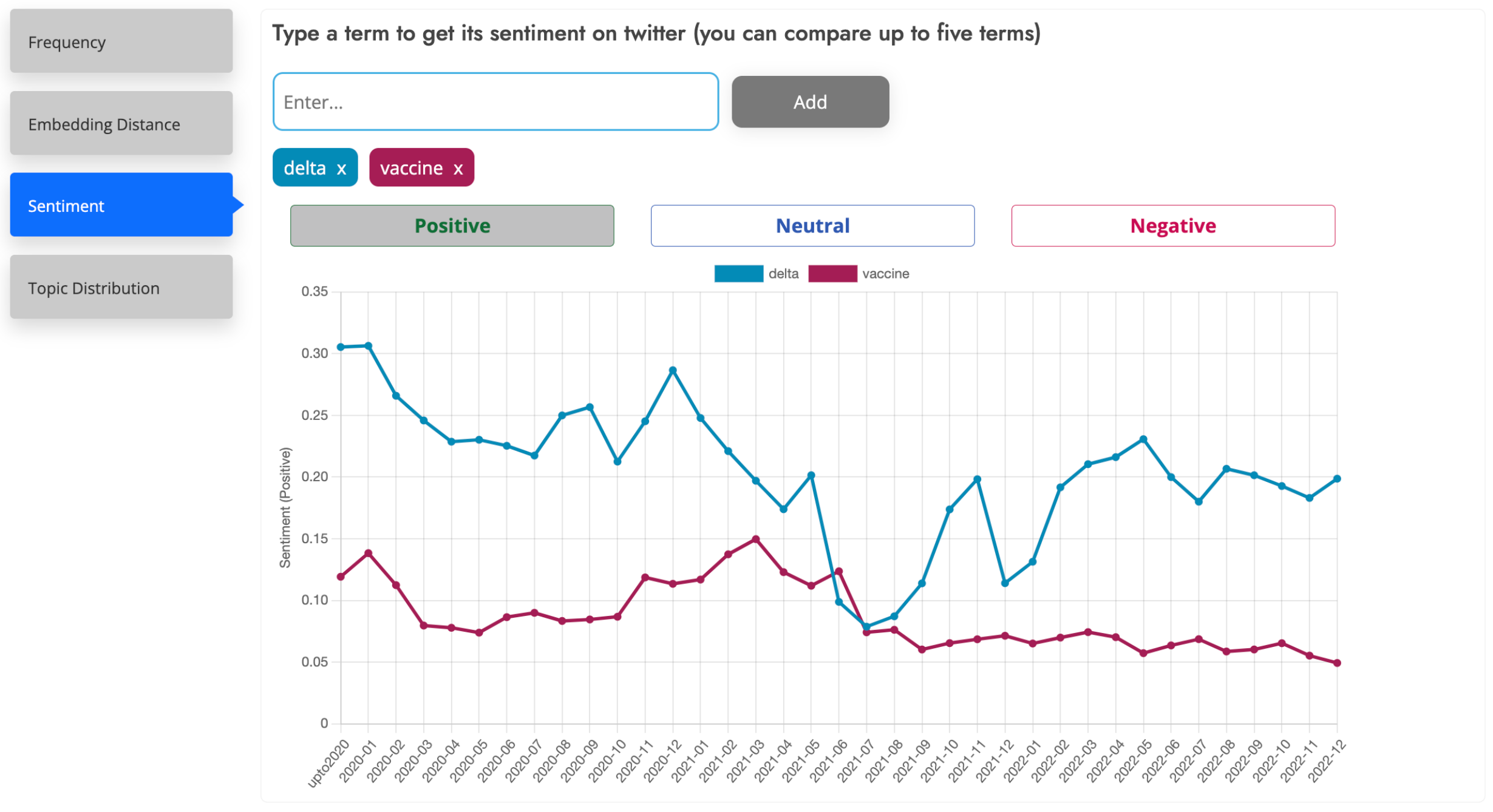}
    \caption{Example of the \textsc{Tweet Insights} demo visualization interface. In this screenshot, the positive sentiment for the terms \textit{delta} and \textit{vaccine} is displayed.}
    \label{fig:demo}
\end{figure*}

\subsubsection{Topic distribution}
We obtain topic distributions for every tweet in our corpus using the same TimeLMs model used for sentiment prediction, but fine-tuned on the multi-label topic classification dataset of \citet{antypas-etal-2022-twitter}\footnote{Covers 19 topics: `arts \& culture', `business \& entrepreneurs', `celebrity \& pop culture', `diaries \& daily life', `family', `fashion \& style', `film tv \& video', `fitness \& health', `food \& dining', `gaming', `learning \& educational', `music', `news \& social concern', `other hobbies', `relationships', `science \& technology', `sports', `travel \& adventure', `youth \& student life'.}, again using all data splits.
%This topic classification model produces scores over 19 topics, covering a wide-range of categories from  to `sports', `fashion \& style', among others.
As with sentiment, the time series resulting from topic predictions corresponds to the average of the mean scores for particular n-grams.
For a better visualization experience, our online demo only shows the top four topics for a particular n-gram, determined from the sum of topic distributions over all periods.

\begin{figure*}[htp]
    \centering
    \includegraphics[width=0.9\linewidth]{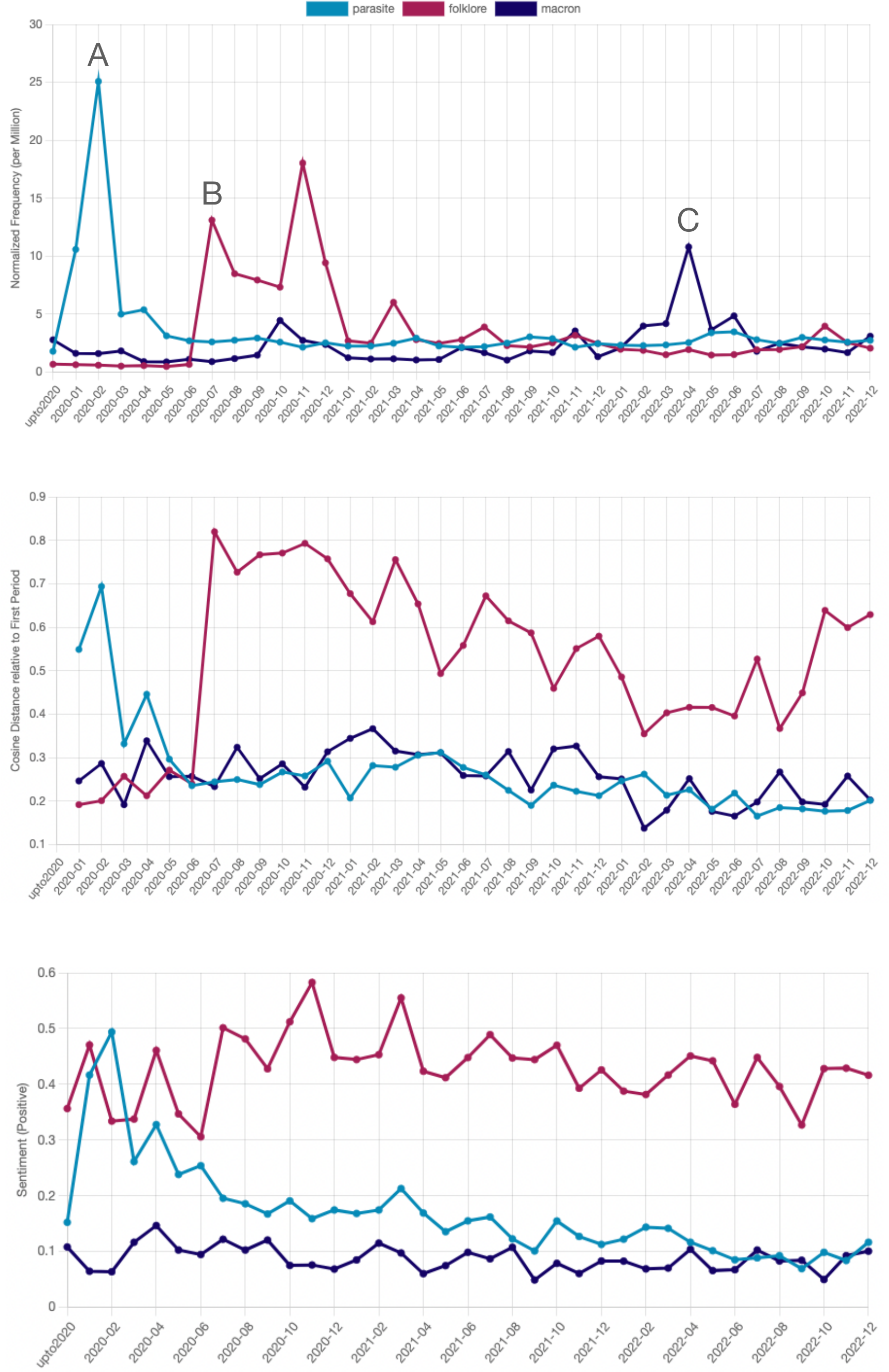}
    \caption{Demo visualizations of our frequency, cosine distance, and positive sentiment time series for the words: \textit{parasite}, \textit{folklore} and \textit{macron}. Note A (2020-02): The movie `Parasite' wins 4 Oscars. Note B (2020-07): Taylor Swift announces her new album named `folklore'. Note C (2022-04): Macron is re-elected as President of France.}
    \label{fig:cases}
\end{figure*}

\begin{figure*}[htp]
    \centering
    \includegraphics[width=0.9\linewidth]{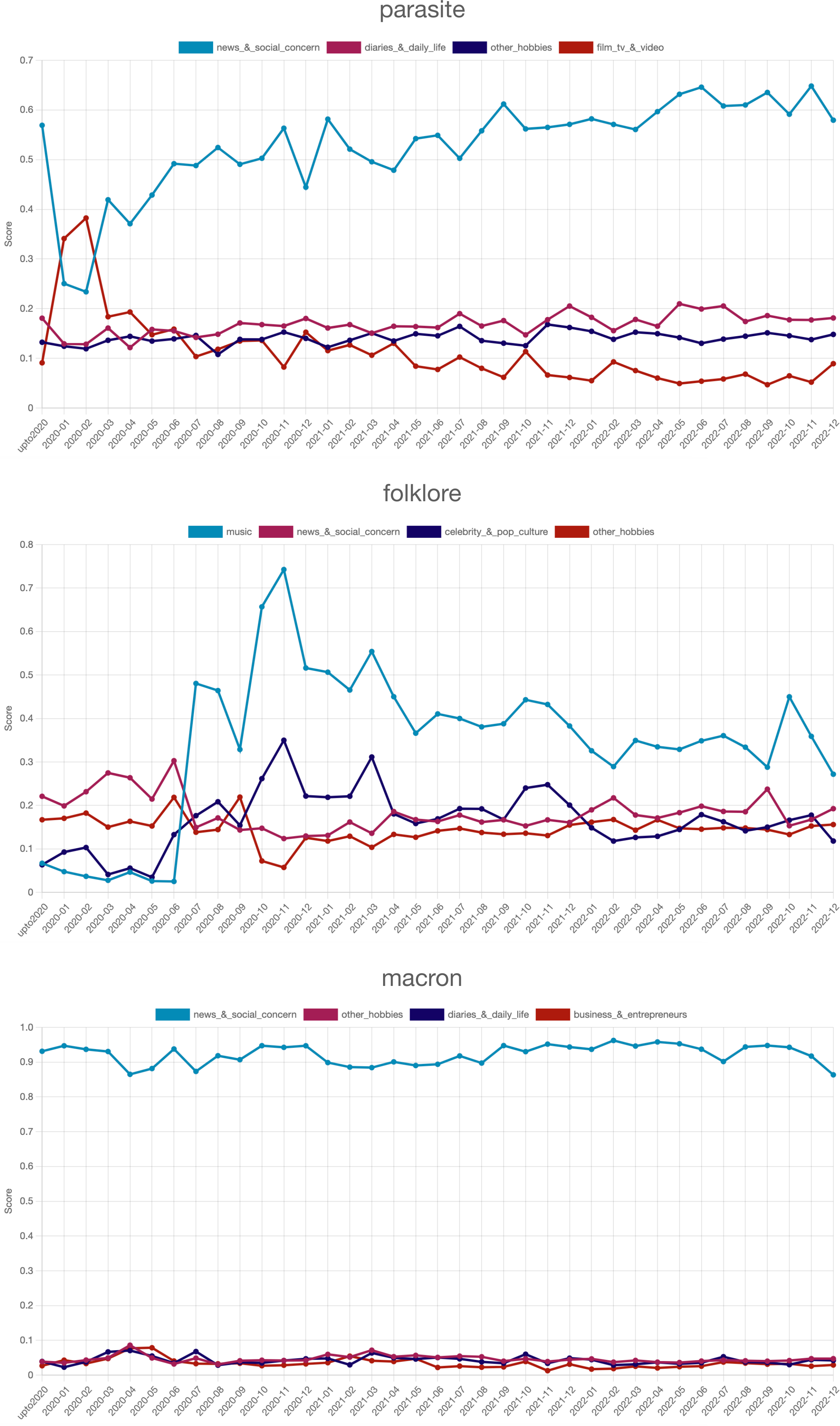}
    \caption{Demo visualization of topic distributions for the words: \textit{parasite}, \textit{folklore}, and \textit{macron} (as in \autoref{fig:cases}). Showing top 4 scoring topics for each word.}
    \label{fig:topics}
\end{figure*}

%\section{Case Studies}
\section{{\textsc{Tweet Insights} Demo}}
\label{sec:demo}

Figure \ref{fig:demo} shows an example of the demo visualization interface. Based on our collected vocabulary and data, as described in the previous section, our demo provides a temporal line plot visualisations from January 2020 onward at monthly intervals. For frequency, similarity, and sentiment visualization, our demo includes the option to visualize multiple words in the same plot. \textsc{Tweet Insights} demo is available at \url{https://tweetnlp.org/insights}.

In this section, we also explore a few case studies using our proposed time series visualizations. As our case studies, we include visualizations for three words (\textit{parasite}, \textit{folklore} and \textit{macron}) in \autoref{fig:cases} and \autoref{fig:topics}, which have been selected to highlight different insights made available by our resource.
Below we describe the some interesting aspects of these cases in detail.

\paragraph{Case 1: \textit{parasite}.}
The frequency plot in \autoref{fig:cases} shows a large spike in frequency for this word around 2020-02, when a movie of the same name won 4 Oscars.
This peak was not sustained for long, and three months later, frequency returned to values much closer to the prior of `up to 2020'.
The connection between this popularity peak and the movie is corroborated by a similarly timed peak in the cosine distance plot, indicating a divergence of the word's associations during this period in relation to its more typical associations.
Interestingly, while the frequency in 2020-02 is double that of 2020-01, the cosine distance plot shows that in 2020-01 its meaning associations had already clearly shifted.
The sentiment plot shows the percentage of tweets that were classified as positive by the model. From it, we can see that the percentage of positive sentiment increased during the awards, but then quickly returned to lower sentiment scores identical to its prior.
Given the decrease in cosine distance closely following that drop in positive sentiment, we can attribute the loss to the word's return to its typical associations (i.e., disease).
Finally, the topic plot in \autoref{fig:topics} further confirms that the movie awards are responsible for the frequency and positivity peak, showing a topic related to `film tv \& video' briefly overtaking the word's typical `news \& social concern' topic.
These results, in combination, provide evidence in support of meaning shift for the word `parasite' during its trending period.

\paragraph{Case 2: \textit{folklore}.}
Similarly to the previous case study, this word also exhibits a relatively short period of increased popularity followed by a return to frequencies close to its prior.
However, in this case, we can see that the cosine distance plot shows a sustained divergence in relation to its prior.
Additionally, and most clearly, its corresponding topic plot in \autoref{fig:topics} shows the `music' topic becoming dominant after both the rise and fall of its popularity peak.
These results suggest that even though the event leading to the increased frequency of this word didn't sustain its popularity, it did lead to a sustained meaning shift, potentially resulting in significantly different word meaning distributions in tweets containing this word sampled before and after the trending event.

\paragraph{Case 3: \textit{macron}.}
We include this word as a case study to show that \textsc{Tweet Insights} also allows for determining whether a popularity peak isn't associated to meaning shift.
The frequency plot shows that this word peaked during the 2022 French Presidential Election. However, the other more semantically relevant plots do not show significant change.

\section{Conclusion}
In this paper, we presented \textsc{Tweet Insights}, a tool to display temporal tendencies on Twitter at the word level considering four distinct but complementary perspectives: frequency, embedding similarity, sentiment, and topic distribution. To calculate all these signals on a large Twitter corpus, we use word embeddings and language models specialized on social media. Through our tool, users can analyse trends more deeply than using Google Trends or alternative frequency-based solutions, benefiting from recent advances in state-of-the-art NLP methods.
While this work focuses on using our proposed time series to detect meaning shift associated to trending events, other applications should also benefit from this resource.% (e.g., ??).
%and assess the frequency, sentiment and meaning variability of words over time.

\section*{Limitations}

As far as limitations are concerned, our platform is limited by several factors. First, the selection of the vocabulary was performed by following some statistical methods that rely on co-occurrence information, and therefore some words/entities may be missing. Second, we focus on a specific time period and visualizations are only available for recent periods from 2020. Since the writing of this paper, the Twitter Academic API has been discontinued. We are currently looking at measures to expand the demo to future time periods. Third, the aggregated information may be affected by several factors such as the reliance on specific word embedding methods and fine-tuned language models in the case of sentiment and topic classification, as well as various types of ambiguity. Finally, the platform is currently only available for English. While we plan to extend it for a few more languages in the near future, this requires a non-trivial amount of extra development and testing.

\section*{Ethics Statement}

This paper presents a visualization demo on user-generated data from Twitter. For this, we have followed all current Twitter regulations with respect to gathering and analysing Twitter data. Moreover, we have removed all URLs and non-verified Twitter mentions from tweets prior to the analysis.
%The ethics statement will not count toward the page limit (8 pages for long, 4 pages for short papers).

%\section*{Acknowledgements}
%This 

\end{spacing}

% Entries for the entire Anthology, followed by custom entries
\bibliography{anthology,custom}
\bibliographystyle{acl_natbib}

\clearpage

\appendix

\section{Corpus Temporal Distribution}
\label{sec:appendix}

\begin{figure*}[b]
    \centering
    \includegraphics[width=1.0\linewidth]{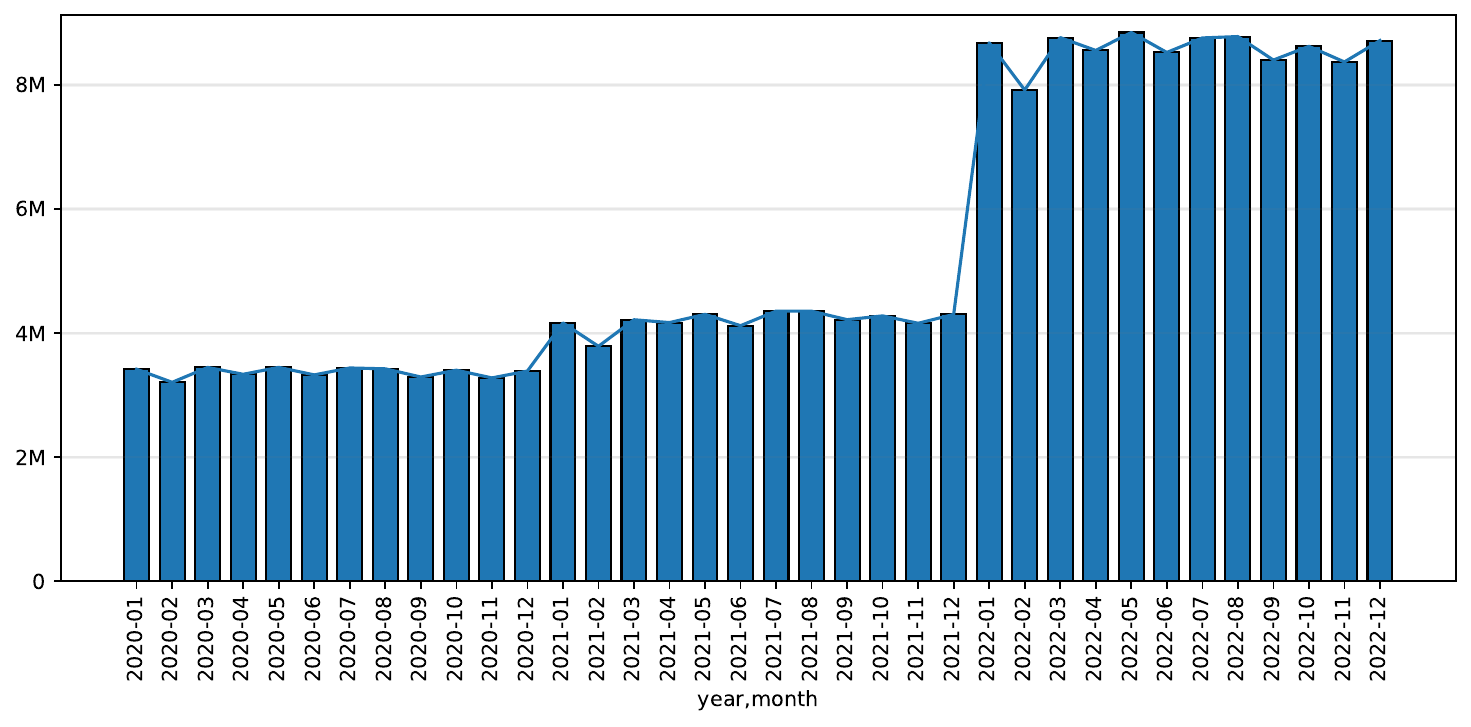}
    \caption{Tweet distribution by year-month intervals. Omitting 27M tweets from the `up to 2020' period.}
    \label{fig:ymdist}
\end{figure*}

\begin{figure*}[b]
    \centering
    \includegraphics[width=1.0\linewidth]{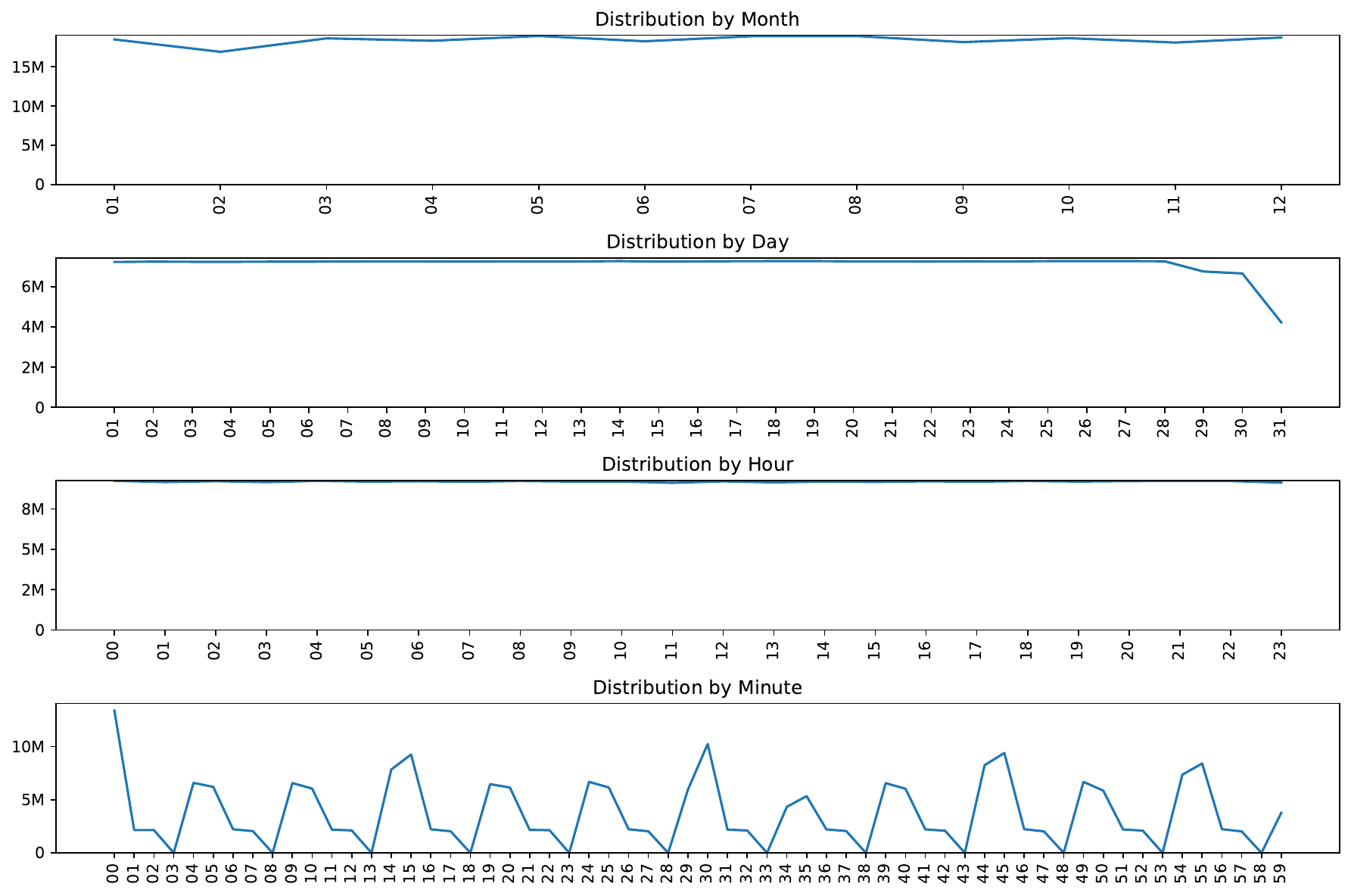}
    \caption{Tweet distribution at different temporal intervals. Highlighting even distribution across different temporal scales, with the exception of the minute-scale distribution (e.g., nearly identical quantity of tweets dated in January and March; no tweets with timestamps at minute 3 of any hour).}
    \label{fig:scalesdist}
\end{figure*}

In this appendix, \autoref{fig:ymdist} and \autoref{fig:scalesdist} demonstrate the similar temporal distribution of tweets underlying our various time series. The distributions are displayed by month, day, hour and minute. 

\end{document}